\documentclass[10pt, a4paper]{article}
\usepackage[]{lrec-coling2024} % this is the new style

\usepackage{graphicx}
\usepackage{tabularx}
\usepackage{soul}
\usepackage{float}
\usepackage{comment}

\usepackage{xcolor}
\usepackage{hyperref}
 \definecolor{darkblue}{rgb}{0, 0, 0.5}
  \hypersetup{colorlinks=true, citecolor=darkblue, linkcolor=darkblue, urlcolor=darkblue}

\usepackage{xstring}
\usepackage{comment}

\usepackage{color}

\usepackage{todonotes}
\usepackage{booktabs}

\definecolor{cadmiumgreen}{rgb}{0.0, 0.42, 0.24}
\definecolor{cadmiumred}{rgb}{0.89, 0.0, 0.13}
\definecolor{cadmiumorange}{rgb}{0.93, 0.53, 0.18}

\title{CLASSLA-web: Comparable Web Corpora of South Slavic Languages Enriched with Linguistic and Genre Annotation}

\name{Nikola Ljubešić, Taja Kuzman} 

\address{Jožef Stefan Institute,  University of Ljubljana, Institute of Contemporary History \\
         Jamova cesta 39, Večna pot 113, Privoz 11, 1000 Ljubljana, Slovenia \\
         nikola.ljubesic@ijs.si, taja.kuzman@ijs.si\\}
\abstract{
This paper presents a collection of highly comparable web corpora of Slovenian, Croatian, Bosnian, Montenegrin, Serbian, Macedonian, and Bulgarian, covering thereby the whole spectrum of official languages in the South Slavic language space. The collection of these corpora comprises a total of 13 billion tokens of texts from 26 million documents. The comparability of the corpora is ensured by a comparable crawling setup and the usage of identical crawling and post-processing technology. All the corpora were linguistically annotated with the state-of-the-art CLASSLA-Stanza linguistic processing pipeline, and enriched with document-level genre information via the Transformer-based multilingual X-GENRE classifier, which further enhances comparability at the level of linguistic annotation and metadata enrichment. The genre-focused analysis of the resulting corpora shows a rather consistent distribution of genres throughout the seven corpora, with variations in the most prominent genre categories being well-explained by the economic strength of each language community. A comparison of the distribution of genre categories across the corpora indicates that web corpora from less developed countries primarily consist of news articles. Conversely, web corpora from economically more developed countries exhibit a smaller proportion of news content, with a greater presence of promotional and opinionated texts.
\\ \newline  }

\begin{document}

\maketitleabstract

%Regular long papers - up to eight (8) pages maximum,* presenting substantial, original, completed, and unpublished work.
%* Excluding any number of additional pages for references, ethical consideration, conflict-of-interest, as well as data and code availability statements.
%Upon acceptance, final versions of long papers will be given one additional page – up to nine (9) pages of content plus unlimited pages for acknowledgments and references – so that reviewers’ comments can be taken into account. For both long and short papers, all figures and tables that are part of the main text must fit within these page limits.

%Papers must be \textbf{anonymized to support double-blind reviewing}. Submissions thus must not include authors’ names and affiliations. The submissions should also avoid links to non-anonymized repositories: the code should be either submitted as supplementary material in the final version of the paper, or as a link to an anonymized repository (e.g., Anonymous GitHub or Anonym Share). Papers that do not conform to these requirements will be rejected without review.

% \citet (cite in text) to get ``author (year)''
%\citep (cite in parentheses) to get ``(author, year)'' citations
%  \citealp (alternative cite without parentheses) to get ``author, year'' citations.
% a language resource should be cited as \citetlanguageresource{Speecon} or \citeplanguageresource{EMILLE}

% Headings should be capitalised in the same way as the main title.
\section{Introduction} 

The South Slavic languages constitute one of the three major branches within the Slavic language family. They are mostly spoken in Central and Southeastern Europe, spanning from the Slovenian Alps, across the Adriatic and Dinarides regions, to Bulgaria and the Black Sea. Despite their widespread use, many languages within this group remain relatively low-resourced and under-represented in the field of natural language processing. According to a recent report on the state of technologies for European languages~\cite{rehm2023european}, the support with the core Language Technologies (LT) -- which include corpora, models and language technologies, such as text and speech processing and machine translation -- was shown to be less than moderate for South Slavic languages, with Bosnian and Macedonian having weak support to no support in all LT areas. In order to facilitate the development of language technologies for these languages, it is crucial to have access to substantial amounts of textual data. Such data serve as the foundation for language technologies, including language models and machine translation systems. Additionally, general text corpora -- that is, corpora that cover broad language use and are not specifically restricted to any particular subject, register or genre -- enable linguists and other researchers to analyze linguistic phenomena based on statistically significant amounts of data. Web crawling, an automated method for collecting text corpora, is one of the primary approaches for quick collection of such data. Recently, the MaCoCu\footnote{\url{https://macocu.eu/}} project~\citep{banon2022macocu} used this method to develop and make freely available monolingual and parallel datasets for over 10 under-resourced languages, including the South Slavic languages.

This paper introduces the CLASSLA-web corpora, which are based on the MaCoCu datasets and have been additionally enriched with linguistic annotation and genre information. This collection of corpora represents, to best of our knowledge, the first comparable corpus collection that encompasses the entire language group. The corpora are considered comparable as they were collected and processed with the same tools and within the same time frame. The paper presents a comprehensive overview of the various stages involved in creating the corpora and offers initial insights into their content based on genre information. The creation of the CLASSLA-web corpora involved three main steps: 1) web crawling based on top-level national domains, and subsequent data post-processing and filtering; 2) linguistic annotation with the latest CLASSLA-Stanza pipeline~\citep{ljubesic-dobrovoljc-2019-neural,tervcon2023classla}, the state-of-the-art pipeline for processing South Slavic languages; and 3) genre annotation, which provides an insight into the functional content of these corpora. Genres are text categories that are defined considering the author's purpose, common function of the text, and the text's conventional form \citep{orlikowski1994genre}. Examples of genres are \textit{News}, \textit{Promotion}, \textit{Legal}, etc.  Genre annotation provides valuable insights into the functional content of the corpora, enabling genre-based corpus studies and more focused natural language processing applications in machine translation~\citep{van2018evaluation} or summarization~\cite{stewart2009genre}.

%To conclude, the main contribution of this paper is a detailed description of the construction of a comparable collection of web corpora that cover an entire language group. The CLASSLA-web corpora provide the largest general corpus for each of the languages covered. Furthermore, these corpora represent one of the first collections of general corpora that are automatically annotated with genre information.

The paper is organized as follows. Firstly, we provide an overview of the related work on construction of web corpora and the automatic genre identification in Section \ref{sec:related}. Next, in Section \ref{sec:corpora-creation}, we present the process of creating and curating the CLASSLA-web corpora, whose content is analyzed in Section \ref{sec:corpora-info}. %To provide insights into the similarities and differences between the corpora, we specifically conduct an analysis of their genre distribution in Section \ref{sec:genre-analysis}. 
Lastly, the paper concludes with Section \ref{sec:conclusion}, where we summarize the main findings and present future work.

\section{Related Work\label{sec:related}}

\subsection{Web Corpora}

The tradition of building web corpora can be followed back to the WaCky initiative~\cite{baroni2009wacky}, inside which first web corpora for large European languages were built. Two other notable initiatives include the CoW corpora~\cite{schafer2012building}, developed for large European languages, and the TenTen corpora~\cite{jakubivcek2013tenten}, created for many languages, including some of South Slavic languages, namely Slovenian\footnote{\url{https://www.sketchengine.eu/sltenten-slovenian-corpus/}} and Bulgarian\footnote{\url{https://www.sketchengine.eu/bgtenten-bulgarian-corpus/}}. However, an important limitation of the TenTen corpora is that they are accessible only through concordancers of the Lexical Computing company, which require a paid subscription.

Recently, numerous web-based datasets have emerged from the Common Crawl\footnote{\url{https://commoncrawl.org}} and the Internet Archive data collections\footnote{\url{https://archive.org}}. Most prominent collections include the cc100 dataset~\cite{conneau-etal-2020-unsupervised}, the mC4 dataset~\cite{xue-etal-2021-mt5}, and the OSCAR dataset~\cite{suarez2019asynchronous}. While these multilingual collections are highly useful for multilingual language modelling tasks, they suffer, inter alia, from blind spots, such as an unexplainable small amount of all Western South Slavic (Slovenian, Croatian, Bosnian, Montenegrin, Serbian) data in OSCAR\footnote{\url{https://oscar-project.github.io/documentation/versions/oscar-2301/}} and an unexpectedly small amount of Croatian content in the mC4 dataset~\cite{xue-etal-2021-mt5}.

The previously developed open web corpora for South Slavic languages adhere to the naming convention of the WaCky~\cite{baroni2009wacky} initiative (e.g., slWaC, hrWaC, srWaC). However, they were actually built using the TenTen technology, more specifically, the SpiderLing crawler~\citep{suchomel2012efficient}, the jusText text extraction tool, and the Onion near-deduplication tool~\citep{pomikalek2011removing}. The first web corpora of South Slavic languages were compiled for Slovenian (slWaC) and Croatian (hrWaC)~\cite{ljubevsic2011hrwac}. Three years later, the list of languages was expanded to include Bosnian (bsWaC) and Serbian (srWaC)~\cite{ljubesic-klubicka-2014-bs}. Additionally, the Croatian and Slovenian crawls were later updated~\cite{erjavec2015slwac}. However, there have been no recent activities related to open web corpora for South Slavic languages.% over the past nine years.

\subsection{Genre Prediction}

Until recently, the available technologies for text classification, including support vector machines and logistic regression classifiers, were insufficient to accurately identify genres across languages and datasets~\citep{sharoff2010web,kuzman2023automatic}. However, recent advancements in deep neural technologies led to a breakthrough in this field. Transformer-based language models~\citep{vaswani2017attention}, specifically those fine-tuned on manually-annotated genre datasets, demonstrated the capability to identify genres in various web corpora and languages (see~\citet{ronnqvist2021multilingual, Kuzman_Ljubesic_Pollak_2022}). The main advantage of Transformer models in this task is that their representations incorporate both lexical and syntactic knowledge about a text, which is crucial for accurate genre prediction~\citep{kuzman2022exploring}. The advancement in these technologies has facilitated the development of multilingual Transformer-based genre classifiers, which can now already be effectively applied to large-scale datasets in various languages. Recently, \citet{laippala2022towards} published the Register Oscar dataset, consisting of 351 million documents in 14 languages to which genre labels were automatically assigned. The dataset was created by applying the multilingual genre classifier \citep{ronnqvist2021multilingual} on the OSCAR datasets \citep{suarez2019asynchronous}. Their classifier was trained on genre datasets in multiple languages and uses a schema with 8 genre labels, such as \textit{Narrative}, \textit{Opinion} and \textit{Interactive Discussion}. The unseen-language classification was evaluated based on annotated datasets in eight new languages (Arabic, Catalan, Chinese, Hindi, Indonesian, Portuguese, Spanish and Urdu). The evaluation results for these new languages demonstrated promising performance, with F1 scores ranging from 0.58 to 0.82.

These endeavors were closely followed by the first experiments with automatic genre annotation on the MaCoCu data. \citet{kuzman2023get} applied the X-GENRE classifier, which is described in more detail in Section \ref{sec:genre-annotation}, to the English part of the parallel MaCoCu corpora. Based on a preliminary manual evaluation, the genre classifier achieved a macro F1 score of 0.73 and a micro F1 score of 0.88 on a small sample of English texts derived from the Slovenian-English MaCoCu corpus. %The authors advised that the accuracy can be further improved by disregarding the label when \textit{Other} is predicted, and labels, predicted with low confidence level, obtained from the raw output. With this intervention, they obtained a much higher classifier’s performance, reaching 0.92 in terms of micro and macro F1 score. The study also provided an analysis of corpora based on differences in genre distribution. The results revealed significant differences between the corpora and showed that automatic genre identification provides valuable insights into corpora content and differences.

\section{Construction of the Corpora}
\label{sec:corpora-creation}

\subsection{Collecting and Curating Web Corpora}

The CLASSLA-web corpora are derived from the web crawls that were gathered and curated inside the MaCoCu project\footnote{\url{https://macocu.eu/}}~\citep{banon2022macocu}. Namely, the following datasets were used: Bosnian web corpus MaCoCu-bs 1.0 \citeplanguageresource{macocu-bs}, Bulgarian web corpus MaCoCu-bg 2.0 \citeplanguageresource{macocu-bg},  Croatian web corpus MaCoCu-hr 2.0 \citeplanguageresource{macocu-hr}, Macedonian web corpus MaCoCu-mk 2.0 \citeplanguageresource{macocu-mk}, Montenegrin web corpus MaCoCu-cnr 1.0 \citeplanguageresource{macocu-cnr}, Serbian web corpus MaCoCu-sr 1.0 \citeplanguageresource{macocu-sr}, and Slovenian web corpus MaCoCu-sl 2.0 \citeplanguageresource{macocu-sl}. They comprise 7 South Slavic languages written in Latin and/or Cyrillic script.

The MaCoCu corpora were collected by crawling the national top-level domains, such as, in the case of Slovenian, the Slovenian top-level domain \texttt{.si}. The primary focus of the crawl was on the top-level domain, but texts from generic domains (\texttt{.com}, \texttt{.net} etc.) were also collected if they were 1. either in the list of seed URLs obtained from previous crawls, or 2. connected well through hyperlinks with the websites in the national top-level domain, and containing sufficient data in the target language. The MaCoCu crawler\footnote{\url{https://github.com/macocu/MaCoCu-crawler}}, which is based on the SpiderLing crawler~\citep{suchomel2012efficient}, was used for the crawling. The crawling process was followed by a thorough post-processing to assure high quality data.  Firstly, the jusText\footnote{\url{http://corpus.tools/wiki/Justext}} tool was employed to remove boilerplate content~\citep{pomikalek2011removing}. Secondly, the onion\footnote{\url{http://corpus.tools/wiki/Onion}} tool was used to identify near-duplicate documents \citep{pomikalek2011removing}. Thirdly, the Monotextor\footnote{\url{https://github.com/bitextor/monotextor/releases/tag/v1.1}} tool was used to enhance language identification accuracy and to evaluate the fluency of paragraphs using a language model. Lastly, we further refined the corpora by removing very short texts (shorter than 75 words or consisting only of paragraphs shorter than 70 characters) and texts originating from web domains that had been manually or automatically identified as automatically-generated\footnote{Further details regarding the preparation of corpora %and the identification of domains classified as ``bad'' 
can be found at \url{https://github.com/macocu/Monolingual-Curation/}.}.

Language identification for certain under-resourced languages within the MaCoCu corpora has proven to be a challenge, as these languages are frequently under-represented in the training data of wide-coverage language identification tools. This challenge is particularly pronounced when attempting to differentiate between Bosnian, Croatian, Montenegrin, and Serbian, as these South Slavic languages exhibit a significant level of mutual intelligibility. Thus, to ensure a high level of accuracy, we employed multiple language identification tools at various stages of the pipeline. Firstly, the identification process began with the Google's Compact Language Detector 2 (CLD2)\footnote{\url{https://github.com/CLD2Owners/cld2}} %~\citep{lui2014accurate} 
 at the document level during the crawling phase. Secondly, the FastSpell\footnote{\url{https://github.com/mbanon/fastspell}} tool was used for a more refined language identification at the paragraph level during the post-processing of the corpora with the Monotextor tool. In these initial two steps, Bosnian, Croatian, Serbian and Montenegrin languages were treated as a single macro-language, and the objective in the case of these languages was to determine whether the text belongs to this macro-language or not. Finally, a specialized classifier was employed to distinguish between these four languages. Specifically, they were identified using a Naive Bayes classifier~\citep{rupnik2023benchic}, which relied on lists of words specific to each variety, which were extracted from web texts published on national top-level domains. This approach assumed that, for instance, the Croatian top-level domain (\texttt{.hr}) is predominantly associated with the Croatian language.

%The MaCoCu project not only ensured the collection of high-quality data but also placed emphasis on enhancing the corpora with comprehensive metadata. Inter alia, the web texts are enriched with information on the URL of the text, personal information and the quality of paragraphs. The latter was determined by labels such as ``short'' or ``good'', which are assigned with the jusText \citep{pomikalek2011removing} tool based on factors such as paragraph length, URL, and stopword density.
%The final corpora can be accessed on the CLARIN.SI repository\footnote{\url{https://www.clarin.si/repository/xmlui/discover}} and can be parsed using the Python module prevert\footnote{\url{https://pypi.org/project/prevert/}}.

\subsection{Genre Annotation}
\label{sec:genre-annotation}

One crucial part of the creation of CLASSLA-web corpora was also enrichment of the corpora with metadata on genres. This information offers valuable insights into the functional content of the corpora. Furthermore, it facilitates the creation of subcorpora based on genres, which can be used for genre-based linguistic analyses. We used the multilingual X-GENRE classifier\footnote{https://huggingface.co/classla/xlm-roberta-base-multilingual-text-genre-classifier}~\citep{kuzman2023automatic} to automatically annotate the corpora with genre labels. The classifier uses the following genre categories: \textit{Information/Explanation}, \textit{Instruction}, \textit{News}, \textit{Legal}, \textit{Promotion},  \textit{Opinion/Argumentation}, \textit{Prose/Lyrical}, \textit{Forum} and \textit{Other} (for a detailed description of labels, refer to~\citet{kuzman2023automatic}). %The classifier is based on the base-size multilingual XLM-RoBERTa Transformer-based model~\citep{conneau-etal-2020-unsupervised}. It was fine-tuned on a combination of three datasets, manually annotated with genre labels: the English CORE~\citep{egbert2015developing}, the English FTD~\citep{sharoff2018functional} and the Slovenian GINCO~\citep{kuzman-rupnik-ljubei:2022:LREC} dataset. Each of the datasets has their own set of categories, which were mapped into a joint schema. The reason for using multiple datasets instead of just one is to assure better generalization of the model to new datasets and languages. 
When tested in the in-dataset scenario, i.e., on the test split that is derived from the same dataset as the training data, the X-GENRE classifier achieves a micro F1 score of 0.80 and a macro F1 score of 0.79. In the cross-dataset scenario, the classifier still maintains a strong performance with a micro F1 score of 0.68 and macro F1 score of 0.69~\citep{kuzman2023automatic}. % A further language-based analysis of the model's performance showed that while performance on Slovenian is lower than on the English instances, it is still satisfactory, with a micro-F1 score of 0.75 and a macro-F1 score of 0.76.

\begin{table}
\centering
\begin{tabular}{lrr}
\hline
\textbf{Language}  & \textbf{Other} & \textbf{Mix} \\ 
\hline
Slovenian (sl) &  2.7\% & 5.0\% \\
Croatian (hr) &  1.9\% & 4.1\% \\
Bosnian (bs) &  1.5\% & 2.9\% \\
Montenegrin (cnr) & 2.0\% & 3.3\% \\
Serbian (sr) & 1.8\% & 3.7\% \\
Macedonian (mk) &  1.2\% & 3.3\% \\
Bulgarian (bg) &  2.1\% & 4.1\% \\
\hline
\end{tabular}
\caption{\label{otherlabel} Percentage of documents annotated with genre labels that do not relate to a specific genre (labels \textit{Other} and \textit{Mix}).
}
\end{table}

We applied the genre classifier to each of the seven CLASSLA-web corpora. The processing was performed on the NVIDIA A100 40GB GPU, with approximately 2,000 predictions executed per second. This resulted in a total processing time of 300 hours.

%It is important to note that genre is a document-level phenomenon and cannot be reliably predicted for very short texts. Thus, following previous work~\citep{kuzman2023get}, texts shorter than 75 words were excluded from annotation and were left without a label. This resulted to approximately 3\% to 4\% of the tokens in each corpus without a genre label. %Consequently, approximately 3\% to 4\% of the tokens in each corpus were deemed too short to be annotated with genre information. 

In addition to genre specific categories, the classifier also returns the category \textit{Other}, which denotes texts that do not fit inside any of the genres present in the schema, as would be the case with an exam, interview, etc. This phenomenon is present in around 2 percent of the documents of each corpus.

Additionally, given that the classifier returns per-class logits that can then be transformed into class probabilities via the softmax function, we introduced a new label, \textit{Mix}. This label is used in cases where none of the categories 
reached a probability of 0.8, which is a relatively infrequent phenomenon, affecting between 3 and 5 percent of the documents in each corpus. We have come to the decision to use the \textit{Mix} label by performing a manual analysis of samples of instances with lower probabilities of the most probable label. The analysis showed that the suggested upper threshold of 0.8 isolates the documents containing multiple genres very well, with both high precision and high recall.

Table~\ref{otherlabel} shows the distribution of documents in texts, annotated as \textit{Other} or \textit{Mix}, therefore the label not having a straightforward application.
%Table~\ref{otherlabel} shows the distribution of tokens in texts, discarded as too short, or annotated as \textit{Other} or \textit{Mix}, therefore the label not having a straightforward application. %The label ``Mix'' implies that the classifier was not certain in its prediction, probably due to the text exhibiting characteristics of multiple genres.  . The proportions of tokens assigned to these three categories (``Short'', ``Other'', and ``Mix'') range from approximately 3\% to 5\%. 
If we omit these labels, we can report that the X-GENRE classifier assigned a specific genre label to approximately 92\%--96\% of documents in each corpus.%85\%--90\% of the tokens in each corpus.

After annotation, we performed a manual analysis of samples from Slovenian, Croatian and Macedonian corpora. Samples consisted of 10 instances per genre label (excluding \textit{Other} and \textit{Mix}), amounting to 80 instances per evaluated corpus. The evaluation showed very high performance of the model, both in Latin and Cyrillic scripts, namely 0.88 in macro F1 for Croatian, 0.93 in macro F1 for Macedonian and 0.94 in macro F1 in case of Slovenian\footnote{More details on the evaluation of the model's cross-lingual performance will be provided in a future publication that we are currently working on.}.

\subsection{Linguistic Annotation}

The final layer of enrichment of the original MaCoCu datasets was linguistic annotation, which enables linguistic analyses and simplified querying of the corpora through concordancers. For linguistic annotation the CLASSLA-Stanza pipeline~\citep{ljubesic-dobrovoljc-2019-neural, tervcon2023classla} was used, which provides the state-of-the-art linguistic annotation of Slovenian, Croatian, Serbian, Bulgarian, and Macedonian.

%Given the high similarity between Bosnian and Montenegrin languages and Serbian, the linguistic annotation was extended to include these languages using the Serbian pipeline.

The CLASSLA-Stanza pipeline is based on the Stanza neural pipeline~\citep{qi2020stanza}, but was further improved for processing of South Slavic languages \citep{tervcon2023classla}. Notable enhancements include the support of external inflectional lexicons, which greatly increases performance for morphologically rich languages~\citep{ljubesic-dobrovoljc-2019-neural}. Additionally, the training datasets used for all models in the pipeline were expanded beyond the Universal Dependencies data. Moreover, the pipeline uses CLARIN.SI-embed word embeddings~\citep{embed.sl, embed.bg, embed.hr, embed.mk, embed.sr} which were trained on larger and more diverse datasets compared to the embeddings used by Stanza. As a result, the CLASSLA-Stanza pipeline demonstrates superior performance compared to Stanza, with error reduction between 34\% and 98\% on the Slovenian official benchmark SloBENCH\footnote{\url{https://slobench.cjvt.si/}}~\citep{slobench} (see~\citet{tervcon2023classla} for further details).

%The CLASSLA-Stanza pipeline initially provided two annotation modules: the standard module for annotating standard texts, and the non-standard module for annotating non-standard texts, such as tweets, blogs, forums, and news comments. However, web corpora typically consist of both types of texts. Thus, a dedicated web module was added to the pipeline \citep{tervcon2023classla}. The module was developed based on the findings of preliminary experiments conducted with the standard and non-standard modules on the Slovenian and Croatian MaCoCu datasets. These experiments demonstrated that the best results for linguistic annotation of web corpora are achieved by using a standard model for tokenization, and non-standard models for all subsequent levels of annotation.
%The analysis showed that for the tokenization, it is advisable to use the standard module, as the non-standard module has a tendency towards producing shorter segments and incorrectly splitting reported speech and sentences with URLs in multiple sentences. The reason behind this is that the non-standard module is optimized for processing social media texts such as tweets. In contrast, the analysis showed that on the other levels of annotation, the non-standard model provides better results. For instance, they handle non-standard word forms quite a bit better than the standard models. Particularly problematic for the standard Slovenian models were forms with missing diacritics, such as ``clovek'' instead of \textit{človek} (human). These were often assigned incorrect lemmas and morphosyntactic tags.

One highly useful feature of the CLASSLA-Stanza pipeline is that it provides non-standard linguistic processing models for Slovenian, Croatian and Serbian. These models are trained on non-standard social media texts, as well as on texts from closely related languages, such as Croatian in the case of the Serbian language, and vice versa. Furthermore, the training data used to train these models had diacritics partially removed, which is a phenomenon often seen in non-curated online texts. The CLASSLA-Stanza system includes a ``web'' processing module that uses a standard tokenizer, but relies on non-standard models for morphosyntactic processing and lemmatization. This setup was shown to be a great solution for linguistic processing of the variety of texts that can be found online. Additionally, the flexibility of these models allows the Croatian web module to be applied to Bosnian and Montenegrin web corpora with high accuracy. Note that Bosnian and Montenegrin language can be considered, in simplified terms, a mixture of the Croatian and Serbian language~\cite{ljubevsic2018borders} and that the Croatian web module in CLASSLA-Stanza is capable of handling features of both.%It is worth noting that Bosnian and Montenegrin language can be considered, in simplified terms, as a mixture of the Croatian and Serbian language~\cite{ljubevsic2018borders}, features of both the Croatian web module in CLASSLA-Stanza is capable of handling.

Thus, we linguistically annotated the CLASSLA-web corpora using the web module of the CLASSLA-Stanza pipeline, which is implemented as a Python library\footnote{\url{https://pypi.org/project/classla/}}. The linguistic processing involved tokenization, morphosyntactic annotation and lemmatization. The CLASSLA-Stanza pipeline also allows annotation on the levels of dependency parsing and named entity recognition. However, these two processing stages are not supported for the Macedonian language due to the lack of required training data for that language. If the research community expresses interest for these two annotation layers in CLASSLA-web corpora for the available languages, or, even more, if appropriate training data for Macedonian are produced in the meantime, we will add these two additional annotation layers in the next version of the CLASSLA-web corpora.  %However, for the Macedonian language, only the first three levels of annotation were applied due to the lack of support for dependency parsing and named-entity recognition for Macedonian in the CLASSLA-Stanza tool. In contrast, the Slovenian corpus was annotated with an additional level of annotation, namely semantic role labeling.
%The final CLASSLA-web corpora have been mounted to concordancers, which are maintained by the CLARIN.SI research infrastructure\footnote{These concordancers are available in four different versions, each offering distinct functionalities, described here: \url{https://www.clarin.si/info/concordances/}.}. The integration of the corpora into the concordancers allows for effortless searching and statistical analysis of data within large text collections, even for individuals with limited technical expertise. As part of the CLASSLA-web release, a tutorial\footnote{\url{https://www.clarin.si/info/k-centre/classla-web-bigger-and-better-web-corpora-for-croatian-serbian-and-slovenian-on-clarin-si-concordancers/}} on the use of corpora on the CLARIN.SI concordancers was provided. It offers guidance on how to search for words, expressions, collocations and analyse frequencies in different genres in CLASSLA-web corpora.
The final CLASSLA-web corpora have been made freely available on the CLARIN.SI repository as:
\begin{itemize}
    \item Bosnian CLASSLA-web.bs corpus~\citep{classla-bs},
    \item Bulgarian CLASSLA-web.bg corpus~\citep{classla-bg},
    \item Montenegrin CLASSLA-web.cnr corpus~\citep{classla-cnr},
    \item Croatian CLASSLA-web.hr corpus~\citep{classla-hr},
    \item Macedonian CLASSLA-web.mk corpus~\citep{classla-mk},
    \item Slovenian CLASSLA-web.sl corpus~\citep{classla-sl},
    \item Serbian CLASSLA-web.sr corpus~\citep{classla-sr}.
\end{itemize}

The corpora are also available on the CLARIN.SI NoSketch Engine concordancers\footnote{\url{https://www.clarin.si/ske/}} which enable easy querying and linguistic analyses.

\section{CLASSLA-web Corpora\label{sec:corpora-info}}

In this section, we perform basic analyses of the seven newly introduced corpora. In the first part, we run some general analyses on the size of each of the corpora in terms of token and document count, as well as some basic analyses of the top-level domains from which the data originate. We proceed with a genre-based analysis of the corpora, exploring the relationship between corpora as well as the relationship of specific genres within them.

\subsection{General Analysis}

\begin{figure*}[!ht]
\begin{center}
%\fbox{\parbox{6cm}{
%This is a figure with a caption.}}
% old picture \includegraphics[scale=0.5]{lrec2020W-image1.eps} 
\includegraphics[scale=0.8]{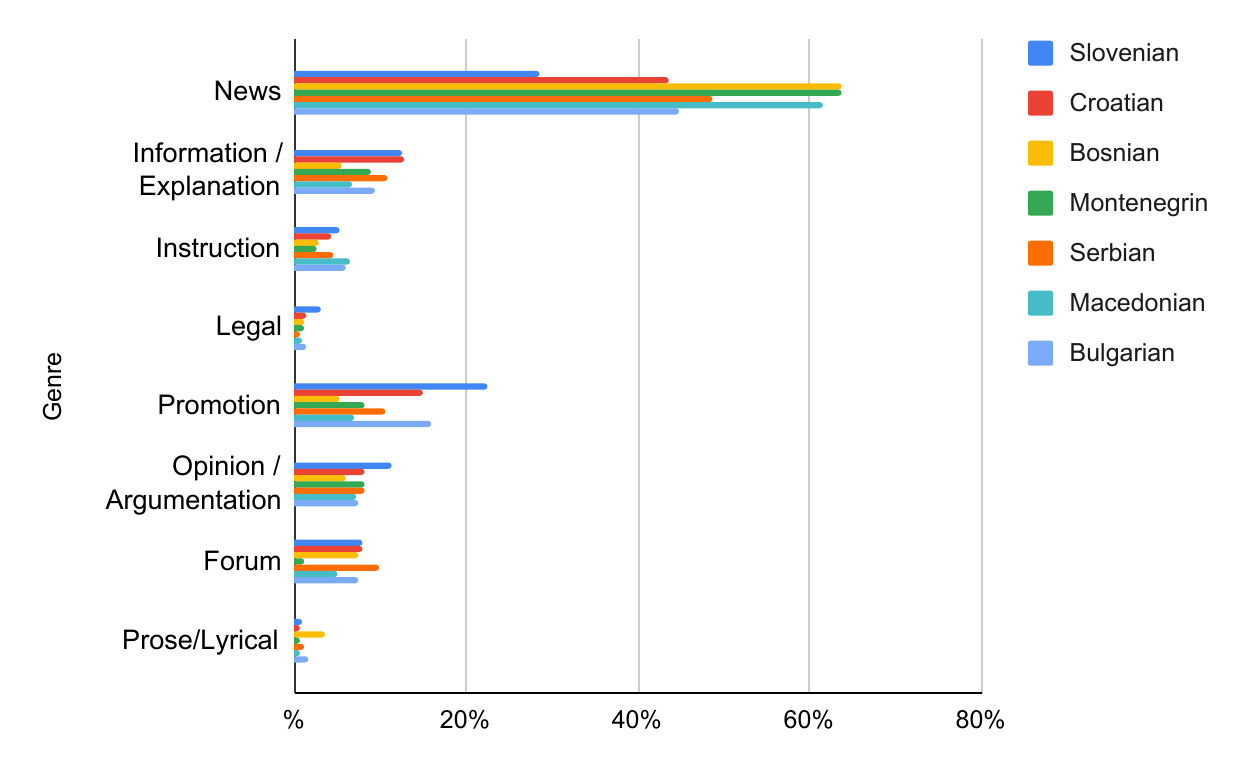} 
\caption{Genre distribution in CLASSLA-web corpora (in percentages of texts).}
\label{fig-genres}
\end{center}
\end{figure*}
% Use chart.png for publishing on ArXiv - otherwise there are errors

The sizes of the resulting corpora for all 7 official South Slavic languages are presented in Table~\ref{corpora-sizes}. The combined corpora amount to almost 13 billion tokens of running text coming from 26 million documents. Among them, the Montenegrin web corpus is the smallest, consisting of 177 million tokens from 401 thousand documents. The largest corpus was derived from the Bulgarian web, consisting of almost 4 billion tokens, obtained from more than 7.4 million documents.

\begin{comment}
\begin{table}
\centering
\begin{tabular}{lrr}
\hline
\textbf{Corpus}      & \textbf{\# tokens} & \textbf{\# docs} \\ 
\hline
CLASSLA-web.sl & 2,282M & 6,302k \\
CLASSLA-web.hr & 2,714M & 8,106k \\
CLASSLA-web.bs & 838M & 2,749k \\
CLASSLA-web.cnr & 186M  & 586k \\
CLASSLA-web.sr & 2,879M & 7,480k \\
CLASSLA-web.mk & 597М  & 1,960k\\
CLASSLA-web.bg & 4,103M & 10,493k \\ 
\hline
\end{tabular}
\caption{\label{corpora-sizes} Sizes of CLASSLA-web corpora in millions of tokens and thousands of documents. The following language codes are used: ``sl'' for Slovenian, ``hr'' for Croatian, ``bs'' for Bosnian, ``cnr'' for Montenegrin, ``sr'' for Serbian, ``mk'' for Macedonian, and ``bg'' for Bulgarian.
}
\end{table}
\end{comment}

\begin{table}
\centering
\begin{tabular}{lrr}
\hline
\textbf{Corpus}      & \textbf{\# tokens} & \textbf{\# docs} \\ 
\hline
CLASSLA-web.sl & 2,153M  & 4,063k  \\
CLASSLA-web.hr & 2,575M  & 5,422k \\
CLASSLA-web.bs & 802M    & 1,993k \\
CLASSLA-web.cnr & 177M    & 401k   \\
CLASSLA-web.sr & 2,765M  & 5,256k  \\
CLASSLA-web.mk & 557M    & 1,482k\\
CLASSLA-web.bg & 3,917M  & 7,456k \\
\hline
Total & 12,948M & 26,076k \\
\hline
\end{tabular}
\caption{\label{corpora-sizes} Sizes of CLASSLA-web corpora in millions of tokens and thousands of documents. The following language codes are used: ``sl'' for Slovenian, ``hr'' for Croatian, ``bs'' for Bosnian, ``cnr'' for Montenegrin, ``sr'' for Serbian, ``mk'' for Macedonian, and ``bg'' for Bulgarian.
}
\end{table}

Each of the presented corpora represents the largest general corpus available for the respective language. What is more, the Macedonian CLASSLA-web corpus is the first linguistically-annotated general corpus of Macedonian. This achievement was made possible not only through the crawl performed inside the MaCoCu project, but especially due to the recent inclusion of basic linguistic processing of Macedonian into the CLASSLA-Stanza linguistic processing pipeline~\cite{tervcon2023classla}.

Table \ref{corpora} shows the differences in the distribution of texts that were derived from the national top-level domain, e.g., \texttt{.si}, and all other, generic domains, such as \texttt{.com}, \texttt{.net} etc.  The percentage of texts originating from the national top-level domain varies significantly, ranging from only 47\% of texts for Montenegrin to 95\% for Macedonian. On average, approximately 70\% of texts were sourced from the national top-level domain, highlighting the importance of crawling beyond the national top-level domain.

\begin{table}
\centering
\begin{tabular}{lrr}
\hline
\textbf{Corpus}   & \textbf{National} & \textbf{Generic} \\ 
\hline
CLASSLA-web.sl &  78.2\% & 21.8\%\\
CLASSLA-web.hr &  75.8\% & 24.2\% \\
CLASSLA-web.bs & 62.5\% & 37.5\% \\
CLASSLA-web.cnr & 46.6\% & 53.4\% \\
CLASSLA-web.sr & 63.6\% & 36.4\% \\
CLASSLA-web.mk & 95.2\% & 4.8\%  \\
CLASSLA-web.bg & 71.1\% & 28.9\% \\
\hline
\end{tabular}
\caption{\label{corpora} Distribution of texts, derived from national top-level domains and other, generic domains.
}
\end{table}

\subsection{Genre-Based Analysis}
\label{sec:genre-analysis}

We start our genre-based analysis of our seven new corpora by plotting the genre distribution in texts across these corpora in Figure~\ref{fig-genres}. %to easier inspect for similarities and differences in prevalence of specific genres.
The first observation to be made is that genres in general are similarly distributed across all seven corpora. The most prevalent genre categories in the analyzed web corpora are \textit{News}, \textit{Information/Explanation}, \textit{Promotion}, \textit{Opinion/Argumentation} and \textit{Forum}. Conversely, \textit{Instruction}, \textit{Legal} and \textit{Prose/Lyrical} are the least represented, with \textit{Prose/Lyrical} accounting for 1\% to 4\% of texts in the corpora. 

On the other hand, the most significant disparities in genre distribution across corpora are observed in the case of \textit{Promotion} and \textit{News}. While the Slovenian web corpus consists only of 28\% of \textit{News} texts but 23\% of \textit{Promotion} texts, two-thirds of the Bosnian, Montenegrin and Macedonian webs are dominated by the \textit{News} genre, with \textit{Promotion} content constituting only 5--8\% of the total texts.

The most significant differences in the occurrence of \textit{Promotion} in web corpora appear to be observed among corpora derived from webs of countries with the greatest disparity in development levels, as measured by gross domestic product (GDP) per capita. Among the seven South Slavic nations, Slovenia has the highest GDP per capita, corrected for purchasing power parity (GDP PPP per capita), whereas the GDP PPP per capita values of Bosnia and Herzegovina, Montenegro and Macedonia are almost half of that of Slovenia.
%Given our background knowledge on the level of development inside the 7 countries, and the fact that Slovenia is the most developed country based on the gross domestic product (GDP) per capita, while Bosnia and Macedonia are the least, Montenegro being on the lower side of the spectrum as well
Given this background knowledge, we decided to examine the relationship between the genre distributions across the 7 web corpora, and the GDP (PPP) per capita for each of the corresponding nations, which serves as a strong metric of development. We inspect the relationship based on the Pearson correlation test. Table~\ref{tab:genre-gdp} shows the results of the correlation test that are statistically significant with p-value below the 0.05 threshold. As anticipated, the \textit{Promotion} genre exhibits the strongest positive correlation with the GDP PPP per capita, with an exceptionally high correlation coefficient of 0.938. Interestingly, the \textit{Information/Explanation}, \textit{Legal} and \textit{Opinion/Argumentation} also demonstrate a relatively strong correlation ranging from 0.86 to 0.87. In contrast, the prevalence of \textit{News} displays a significant negative correlation with the GDP (PPP) per capita, with a correlation coefficient of -0.90.

\begin{table}
\centering
\begin{tabular}{lll}
\hline
\textbf{Genre}                     & \textbf{Pearson r }    & \textbf{p-value }       \\
\hline
Promotion                 & 0.938     & 0.002**   \\
Opinion   & 0.873     & 0.010*   \\
Information & 0.861     & 0.013*   \\
Legal                     & 0.861     & 0.013*   \\
News                      & -0.900    & 0.006*   \\
%Instruction               & 0.189     & 0.685   \\
%Forum                     & 0.267     & 0.562   \\
%Prose/Lyrical             & -0.427    & 0.339   \\
%Mix                       & \textbf{0.941}     & 0.002**   \\
%Other                     & \textbf{0.863}     & 0.012*  \\
\hline
\end{tabular}
\caption{\label{tab:genre-gdp} Results of the Pearson correlation test of the GDP (PPP) per capita for each South Slavic country and the distribution of each genre in CLASSLA-web corpora. Asterisks denote p-values: ** for p<0.005 and *~for p<0.05. Only correlations that are statistically significant, i.e., with the p-value below 0.05, are included. \textit{Information/Explanation} is shortened to Information, and \textit{Opinion/Argumentation} is shortened to Opinion.
}
\end{table}

Given the observation that \textit{News} appears to have a negative correlation with the other genres, as evidenced by the contrasting shapes in Figure~\ref{fig-genres}, we also conducted a pairwise Pearson correlation test among all the genres. The results, presented in Table~\ref{tab:genre-correlation}, reveal a nearly perfect negative correlation of -0.972 between the relative frequency of \textit{News} and \textit{Promotion}. This suggests that, as the country evolves from less-developed to better-developed, there is a phenomenon wherein newspaper content in the country's web is increasingly replaced by promotional material. Additionally, high, albeit lower, correlation coefficients can be observed between the \textit{News} genre on one side and the \textit{Information/Explanation}, \textit{Opinion/Argumentation} and \textit{Legal} genre. This analysis allows us to classify the genres into two distinct clusters -- one consisting of the \textit{News} content, and the other consisting of \textit{Information/Explanation}, \textit{Opinion/Argumentation}, \textit{Promotion} and \textit{Legal} texts. Based on these findings, we can suggest a preliminary hypothesis that, as a country experiences economic development, its web becomes more diverse and incorporates a greater variety of promotional, opinionated, informational, and legal content. It is important to note that this observation does not take into account other possible factors that might have an impact on the differences between South Slavic webs. Despite being very preliminary, we put forward this hypothesis in hopes of sparking interest for further research on this phenomenon in the wider research community. %which we intend to further investigate through a comprehensive review of relevant literature and rigorous testing in future research.

\begin{table}
\centering
\begin{tabular}{lll}
\hline
\textbf{Genre pair}                & \textbf{Pearson r }    & \textbf{p-value }       \\
\hline
%News                    & Mix                     & -0.979                                  & 0.000**                       \\
News, Promotion               & -0.972                                  & 0.000**                       \\
News, Information & -0.833                                  & 0.020*                       \\
%News                    & Other                   & -0.817                                  & 0.025*                       \\
News, Opinion   & -0.812                                  & 0.026*                       \\
News, Legal                   & -0.777                                  & 0.040*                       \\
Information, Opinion   & 0.783                                   & 0.037*                       \\
Information, Promotion               & 0.813                                   & 0.026*                       \\
%Legal                   & Mix                     & 0.814                                   & 0.026*                       \\
%Opinion/Argumentation   & Other                   & 0.827                                   & 0.022*                       \\
Promotion, Opinion   & 0.831                                   & 0.021*                       \\
Legal, Promotion               & 0.834                                   & 0.020*                       \\
%Mix                     & Other                   & 0.841                                   & 0.018*                       \\
%Information/Explanation & Mix                     & 0.849                                   & 0.016*                       \\
Legal, Opinion   & 0.851                                   & 0.015*                       \\
%Legal                   & Other                   & 0.852                                   & 0.015*                       \\
%Opinion/Argumentation   & Mix                     & 0.869                                   & 0.011*                       \\
%Promotion               & Other                   & 0.882                                   & 0.009*                       \\
%Promotion               & Mix                     & 0.990                                   & 0.000**      \\                      
\hline
\end{tabular}
\caption{\label{tab:genre-correlation} Results of the Pearson correlation test over pairwise genre categories. Categories \textit{Mix} and \textit{Other} are not included in the analysis. Only correlations that are statistically significant, i.e., with the p-value below 0.05, are included. Asterisks denote p-values: ** for p<0.005 and *~for p<0.05. \textit{Information/Explanation} is shortened to Information, and \textit{Opinion/Argumentation} is shortened to Opinion.
}
\end{table}

\begin{comment}
\begin{table*}
\centering
\begin{tabular}{llllllll}
\hline
Genre/Dataset           & Slovenian & Croatian & Serbian & Bosnian & Montenegrin & Macedonian & Bulgarian \\
News                    & 28.45     & 43.62    & 48.78   & 63.75   & 63.80       & 61.65      & 44.76     \\
Information/Explanation & 12.58     & 12.83    & 10.76   & 5.47    & 8.98        & 6.72       & 9.35      \\
Instruction             & 5.34      & 4.27     & 4.56    & 2.85    & 2.69        & 6.58       & 6.02      \\
Legal                   & 3.07      & 1.33     & 0.77    & 1.11    & 1.17        & 0.85       & 1.33      \\
Promotion               & 22.60     & 14.94    & 10.52   & 5.36    & 8.16        & 6.94       & 16.00     \\
Opinion/Argumentation   & 11.35     & 8.29     & 8.18    & 6.10    & 8.12        & 7.27       & 7.38      \\
Forum                   & 8.04      & 8.06     & 9.87    & 7.39    & 1.05        & 4.95       & 7.40      \\
Prose/Lyrical           & 0.87      & 0.60     & 1.08    & 3.55    & 0.76        & 0.58       & 1.61      \\
Mix                     & 4.97      & 4.14     & 3.69    & 2.91    & 3.29        & 3.31       & 4.07      \\
Other                   & 2.74      & 1.91     & 1.78    & 1.51    & 1.97        & 1.15       & 2.09     \\
\hline
\end{tabular}
\caption{\label{table-genres} Genre distribution in CLASSLA-web corpora (in percentages).
}
\end{table*}
\end{comment}

\section{Conclusions}
\label{sec:conclusion}

This paper introduces a collection of comparable web corpora for South Slavic languages. To the best of our knowledge, this is the first corpus collection that comprises comparable general corpora for all languages within a language group. Furthermore, these corpora represent the largest general corpora available for each respective language. The corpus collection comprises 13 billion tokens and 26 million documents in total. Additionally, for the least resourced language in this group, the Macedonian language, the Macedonian CLASSLA-web corpus is the first general linguistically-annotated corpus.

The creation of these corpora was made possible through the contribution of multiple separate endeavors. This paper aims to document all the necessary steps that were undertaken in order to create the corpora. Firstly, the MaCoCu project~\citep{banon2022macocu} played a crucial role by collecting the corpora based on a comprehensive crawl of the national top-level domains and well-interconnected general domains. Additionally, the project implemented various post-processing methods to ensure the high quality of the final datasets. Notably, that research highlighted the significant challenges associated with language identification of less prevalent languages, especially between closely-related South Slavic languages, such as the mutually intelligible Croatian, Bosnian, Montenegrin and Serbian languages.% In this paper, we describe how three distinct language identification tools were employed to ensure accurate identification between Serbian, Croatian, Bosnian, and Montenegrin languages, which are closely-related and mutually intelligible.

Secondly, high-quality linguistic annotation of the CLASSLA-web corpora was made possible by recent improvements of the CLASSLA-Stanza pipeline \citep{tervcon2023classla}. In addition to improving annotation accuracy with extended training datasets and embeddings, the new version of CLASSLA-Stanza pipeline provides a ``web'' module, specifically tailored for linguistic annotation of web corpora. This is particularly advantageous, as web corpora pose a unique challenge to linguistic processing pipelines due to their composition of standard and non-standard texts. What is even more, while Bosnian and Montenegrin do exhibit combinations of linguistic features that distinguish them from Croatian and Serbian, most of these features are covered either in the one or the other language training data, enabling for the Croatian ``web'' module, trained on standard and non-standard datasets of both Croatian and Serbian, to very successfully annotate the Bosnian and Montenegrin CLASSLA-web corpora.  Linguistic annotation facilitates comprehensive analyses of language phenomena for corpus linguists, who perform analyses based on part-of-speech and other linguistic information.

Thirdly, we applied to the corpora a recently-developed multilingual genre classifier. This automatic annotation with genre categories allowed us to enrich the datasets with valuable information on the communicative function of documents inside the seven corpora. Our findings demonstrate that the CLASSLA-web corpora exhibit similar genre distributions. However, we observed that certain corpora, such as Bosnian, Montenegrin, and Macedonian, predominantly consist of news content. In contrast, the Slovenian corpus contains a smaller proportion of news content, with promotional texts representing a significant portion of the corpus. %Based on these initial observations, we examined the relationship between the economic development of the countries where the target language of a corpus is predominantly spoken, and the distribution of genres inside the respective corpus. Our analysis revealed that web corpora from countries with higher GDP (PPP) per capita tend to have a greater amount of promotional content and a lower proportion of news content. In future research, we intend to expand upon these observations by conducting a comprehensive literature review and more extensive experiments, covering also old versions of web corpora available for some of the languages.
In future research, we plan to significantly extend explorations of genres inside the CLASSLA-web corpora. Firstly, we plan to perform manual analysis on all 7 languages to ascertain what is the overall performance of the multilingual genre classifier on each of the corpora. Secondly,  we will investigate possible patterns of biases that could negatively impact downstream research relying on genre labels obtained from multilingual models. Finally, we plan to use genre information to analyze the linguistic characteristics of genres in South Slavic web corpora.

Given the substantial sizes of the CLASSLA-web corpora, they are immensely valuable for the development of language technologies for South Slavic languages and future linguistic analyses. The corpora are already being used for the development of BERT-like and generative pretrained (GPT) language models\footnote{See for instance the XLM-R-BERTić (\url{https://huggingface.co/classla/xlm-r-bertic}) and YugoGPT (\url{https://huggingface.co/gordicaleksa/YugoGPT}) models.}, specific to South Slavic languages. Furthermore, the data can be useful as the starting point for development of manually-annotated training data for numerous tasks. For instance, as part of the Slovenian EMMA project\footnote{\url{https://emma.ijs.si/en/about-project/}}, focused on providing NLP solutions to the media industry, samples of the texts annotated as \textit{News} are planned to be used for development of datasets for multilingual topic prediction in news. Moreover, the corpora are highly useful for linguistic analyses as was already shown by their predecessors slWaC, hrWaC~\cite{ljubevsic2011hrwac}, srWaC~\cite{ljubesic-klubicka-2014-bs} and others. Accordingly, we have made the corpora available through the CLARIN.SI concordancers\footnote{\url{https://www.clarin.si/ske/}} as well, to enable easy corpus querying. To promote their use in the wider linguistic community, which also includes language teachers and digital humanists, the use of the corpora through concordancers will be presented in CLASSLA Express\footnote{\url{https://www.clarin.si/info/k-centre/workshops/classla-express/}}, a series of five workshops that will take place in four South Slavic countries.

The MaCoCu approach demonstrated significant success in automatically collecting texts by crawling the top-level domains and beyond, resulting in the creation of the largest general text collections for each of the targeted South Slavic languages. Further enrichment of these text collections with linguistic and genre information resulted in a corpus collection that was not just collected, but also enriched in a highly comparable way, enabling comparable insights in the functional composition of these corpora, as well as unlocking the linguistic research potential of the corpora by performing multi-layer linguistic annotation. To ensure transparency and reproducibility, all steps involved in the process are publicly available. Our future plans involve conducting iterative crawling, following the MaCoCu web corpora collection approach, and enrichment of the South Slavic languages on a yearly or bi-yearly basis. We have set up a crawling infrastructure inside the CLARIN.SI research infrastructure\footnote{\url{https://www.clarin.si/}} which is dedicated to iterative crawling of South Slavic webs and webs in other languages. We have already started performing a new run of crawling for Slovenian, Croatian, Serbian, Bosnian and Montenegrin, which will be followed by crawling of Bulgarian and Macedonian. This will allow us to further expand and update the current version of the CLASSLA-web corpora. Consistent updating the corpora will also enable research on how the texts and their distribution on the web are evolving through time, and also enable research in the field of semantic change for South Slavic languages. Additionally, by presenting this process in this paper, we aim to inspire similar initiatives to develop web corpora for other languages lacking large high-quality corpora.
%MaCoCU experiments: crawling the same TLD some time after the previous crawl, for which the Icelandic TLD was selected as a use case. They crawled the top national domains in 2021 and in 2023, with 18 months apart. They observed an increment of 37\,\% of total text extracted from the merged data. These results encouraged us to continue repeated crawls of South Slavic corpora, which we plan to do the following year.

\section{Acknowledgments}
The research presented in this paper was conducted within the research project ``Basic Research for the Development of Spoken Language Resources and Speech Technologies for the Slovenian Language'' (J7-4642), the research project ``Embeddings-based techniques for Media Monitoring Applications'' (L2-50070, co-funded by the Kliping d.o.o. agency) and within the research programme ``Language resources and technologies for Slovene'' (P6-0411), all funded by the Slovenian Research and Innovation Agency (ARIS).

This work has received funding from the European Union's Connecting Europe Facility 2014-2020 - CEF Telecom, under Grant Agreement No. INEA/CEF/ICT/A2020/2278341. This communication reflects only the author's view. The Agency is not responsible for any use that may be made of the information it contains.

We would like to thank Petra Bago, Virna Karlić and Lidija Milković in helping to validate and improve the content of the Croatian corpus. We would also like to extend our gratitude to Marija Runić for giving guidance through the complexity of the Bosnian web. We are finally grateful to all our collaborators in the MaCoCu  project who made these corpora significantly better.

\subsection{Ethical Considerations and Limitations}

We are aware that using data that was collected from the web can raise questions of respecting the intellectual property and privacy rights of the original authors of the texts. The authors of the MaCoCu datasets, on which the CLASSLA-web corpora are based on, assured that no sensitive data would be included by only collecting the texts that were freely accessible. Nevertheless, we are aware that the datasets might still include some texts that the authors do not consent to be included. To mitigate this, the CLASSLA-web corpora are published with a notice, which informs the authors of the text that the texts can be taken out of the corpora upon their request. 

%Please note that extra space is allowed after the 8th page (4th page for short papers) for an ethics/broader impact statement and a discussion of limitations. At submission time, this means that if you need extra space for these sections, it should be placed after the conclusion so that it is possible to rapidly check that the rest of the paper still fits in 8 pages (4 pages for short papers). Ethical considerations sections, limitations, acknowledgements, and references do not count against these limits. For camera-ready versions 9 pages of content will be allowed for long (5 for short) papers.

\nocite{*}
\section{Bibliographical References}\label{reference}
%\label{main:ref}

\bibliographystyle{lrec_natbib}
\bibliography{bibliography}

%\section{Language Resource References}
%\label{lr:ref}
%\bibliographystylelanguageresource{lrec_natbib}
%\bibliographylanguageresource{languageresource}

%Appendices or supplementary material will be allowed ONLY in the final, camera-ready version, but not during submission, as papers should be reviewed without the need to refer to any supplementary materials.

%Each \textbf{camera ready} submission can be accompanied by an appendix usually being included in a main PDF paper file, one \texttt{.tgz} or \texttt{.zip} archive  containing software, and one \texttt{.tgz} or \texttt{.zip} archive containing data.
%\appendix
%\section{Appendix}

\end{document}